\documentclass[letterpaper]{article} 
\usepackage{aaai24}  
\usepackage{times}  
\usepackage{helvet}  
\usepackage{courier}  
\usepackage[hyphens]{url}  
\usepackage{graphicx} 
\usepackage{natbib}  
\usepackage{caption} 
\usepackage{algorithm}
\usepackage{algorithmic}
\usepackage{amsfonts}

\usepackage{newfloat}
\usepackage{listings}
\DeclareCaptionStyle{ruled}{labelfont=normalfont,labelsep=colon,strut=off} 
\floatstyle{ruled}
\newfloat{listing}{tb}{lst}{}
\floatname{listing}{Listing}
\title{Unsupervised evaluation of GAN sample quality:\\ Introducing the TTJac Score}
\author{
    Egor Sevriugov\textsuperscript{\rm 1}, 
    Ivan Oseledets\textsuperscript{\rm 1} 
}
\affiliations{
    \textsuperscript{\rm 1}Skolkovo institute of science and technology\\
Moscow, Russia 121205\\

    egor.sevriugov@skoltech.ru, I.Oseledets@skoltech.ru
}

\begin{document}

\maketitle

\begin{abstract}
Evaluation metrics are essential for assessing the performance of generative models in image synthesis. However, existing metrics often involve high memory and time consumption as they compute the distance between generated samples and real data points. In our study, the new evaluation metric called the "TTJac score" is proposed to measure the fidelity of individual synthesized images in a data-free manner. The study first establishes a theoretical approach to directly evaluate the generated sample density. Then, a method incorporating feature extractors and discrete function approximation through tensor train is introduced to effectively assess the quality of generated samples. Furthermore, the study demonstrates that this new metric can be used to improve the fidelity-variability trade-off when applying the truncation trick. The experimental results of applying the proposed metric to StyleGAN 2 and StyleGAN 2 ADA models on FFHQ, AFHQ-Wild, LSUN-Cars, and LSUN-Horse datasets are presented. The code used in this research will be made publicly available online for the research community to access and utilize.
\end{abstract}

\section{Introduction}

Advancements in Generative Adversarial Networks (GANs) \cite{GAN} have led to a wide range of applications, including image manipulation \cite{manipulation4,manipulation5,manipulation6}, domain translation \cite{translation1,translation2,translation3,translation4,translation5}, and image/video generation \cite{ivgeneration1,ivgeneration2,ivgeneration3,ivgeneration4,ivgeneration5}. GANs have demonstrated high-quality results in these tasks, as validated by standard evaluation metrics such as Fréchet inception distance (FID) \cite{fid}, kernel inception distance (KID) \cite{kid}, Precision \cite{precisionrecall}, and Recall \cite{precisionrecall}. These metrics are typically based on clustering real data points using the k-nearest neighbours algorithm. Initially, real images are passed through a feature extractor network to obtain meaningful embeddings, and pairwise distances to other real images are computed for the algorithm. In the evaluation stage, the fidelity of an individual sample is determined by computing its distance to the clusters of real manifold. However, this procedure can be computationally expensive and memory-intensive, particularly for large datasets, as all real embeddings need to be stored.

In order to overcome these challenges, this research introduces a novel metric for evaluating the quality of individual samples. Instead of assessing sample fidelity relative to the real manifold, this metric directly calculates the density of a sample by utilizing only the trained generator. The computation process involves evaluating the model Jacobian, which can be particularly demanding for high-resolution models. To mitigate the memory and time costs associated with this computation, feature extractors are employed to reduce the size of the Jacobian.

Furthermore, the proposed metric function is approximated on a discrete grid using tensor train decomposition. This approximation provides a significant reduction in inference time since the batch of sample scores is only required to compute the tensor decomposition stage. The evaluation procedure simply entails obtaining the value of decomposed tensor at specified indexes.

The proposed metric function also has an application in the sampling procedure, particularly as an enhancement for the truncation trick. The truncation trick \cite{truncationtrick} operates by sorting samples based on the norm of the input vector, which can be effectively replaced by the TTJac score. This upgrade provides a better trade-off between fidelity and variability compared to the standard technique.

\begin{figure*}[t]
\centering
\includegraphics[width=0.9\textwidth]{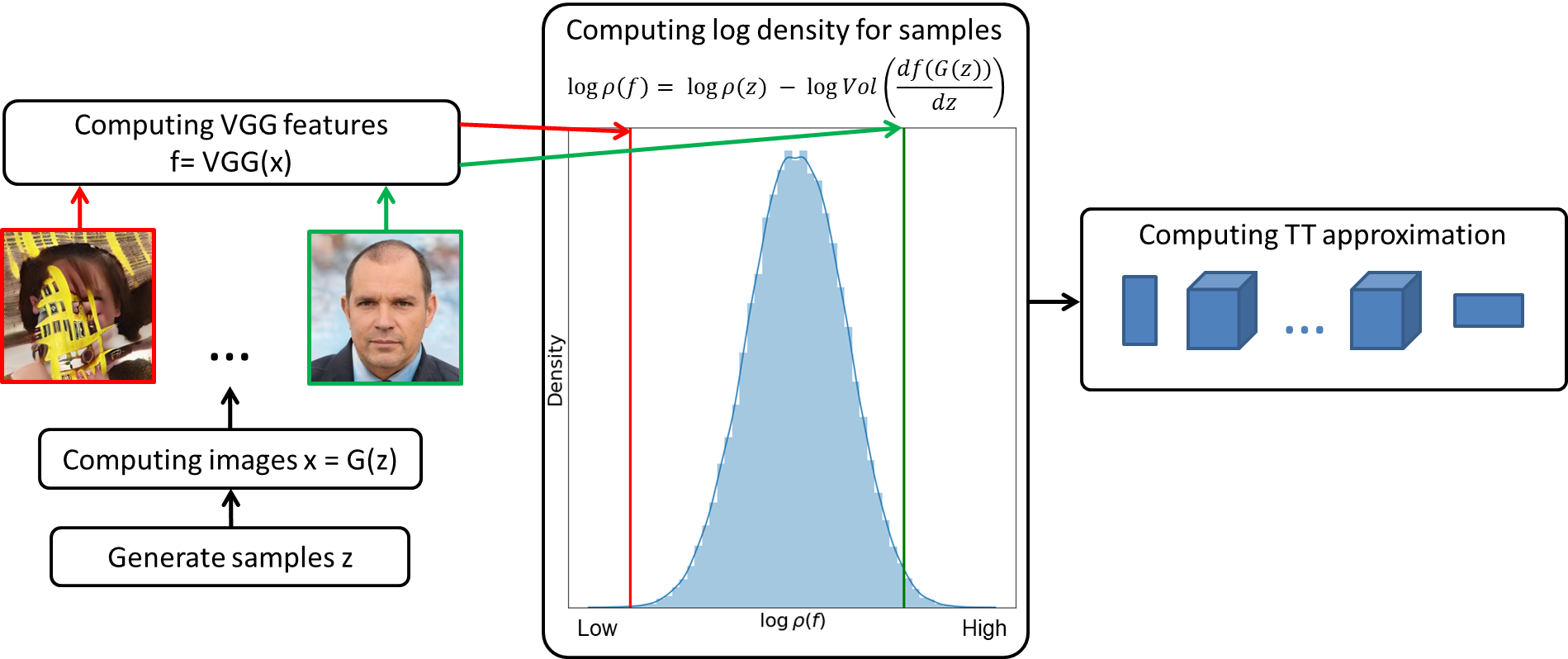}
\caption{The general pipeline of the presented work involves several steps. Firstly, latent code samples $z$ are generated from a normal distribution, then the generated latent codes are passed through the generator network, which produces corresponding images $x$. The VGG feature extractor is employed to extract meaningful features $f$. After obtaining the features, the computation of feature density is carried out using generalized change of variables formula \cite{varchange}. Finally, the metric score samples are approximated using the Tensor Train (TT) algorithm.}
\label{fig:pipeline}
\end{figure*}

To evaluate the effectiveness of the proposed metric function, standard GAN models like StyleGAN 2 \cite{ivgeneration4} and StyleGAN 2 ADA \cite{ivgeneration5} were considered, using various datasets such as Flickr-Faces-HQ Dataset (FFHQ) \cite{ivgeneration3}, AFHQ Wild \cite{translation4}, LSUN Car \cite{lsun}, and LSUN Horse \cite{lsun}. In summary, the contributions of this paper include:

\begin{enumerate}
    \item Introducing a new metric for sample evaluation that does not rely on dataset information.
    \item Presenting a methodology for effective usage of the proposed metric, involving feature extractors and tensor train approximation.
    \item Proposing a metric-based upgrade for the truncation trick, enabling a better trade-off between fidelity and variability.
\end{enumerate}

\section{Method}

The primary component of a GAN model is the generator network, denoted as $G$, which generates an image $x$ from a given latent code (network input) $z$. Typically, the evaluation of individual sample quality $x = G(z)$ is performed using a realism score or trucnation trick.

The realism score requires access to a dataset to compute distances for the k-nearest neighbors algorithm, while truncation trick assesses sample fidelity based on the norm of the corresponding input vector. This approach allows for resampling latent codes that lie outside a chosen radius.

In contrast, we propose a metric function that defines the score based on the density of the generator output. We use the generalized change-of-variable formula \cite{varchange}:

$$\rho(z) = \rho(x)\mathrm{Vol}(J)$$
where $J = dG(z)/dz$ represents the generator Jacobian. By taking the logarithm and making the appropriate substitution, the final expression for the score function is derived:
$$s(x) = \log \rho(x) = \log(\rho(z)) - \log(\mathrm{Vol}(J))$$
where $\mathrm{Vol}(J) = \frac{1}{2}\log(\det(J^TJ)) = \sum\limits_{i=1}^N\log(\sigma_i(J))$, $\sigma_i(J)$ represents the $i$-th singular value of the Jacobian. Finally score function turns into:
$$s(x) = \log(\rho(z)) - \sum\limits_{i=1}^N\log(\sigma_i(J))$$
At this stage, we considered the image density in pixel representation. Nevertheless, the proposed idea can be applied to an arbitrary representation of the image. This aspect is discussed in the next part.

\begin{figure*}[t]
\centering
\includegraphics[width=0.9\textwidth]{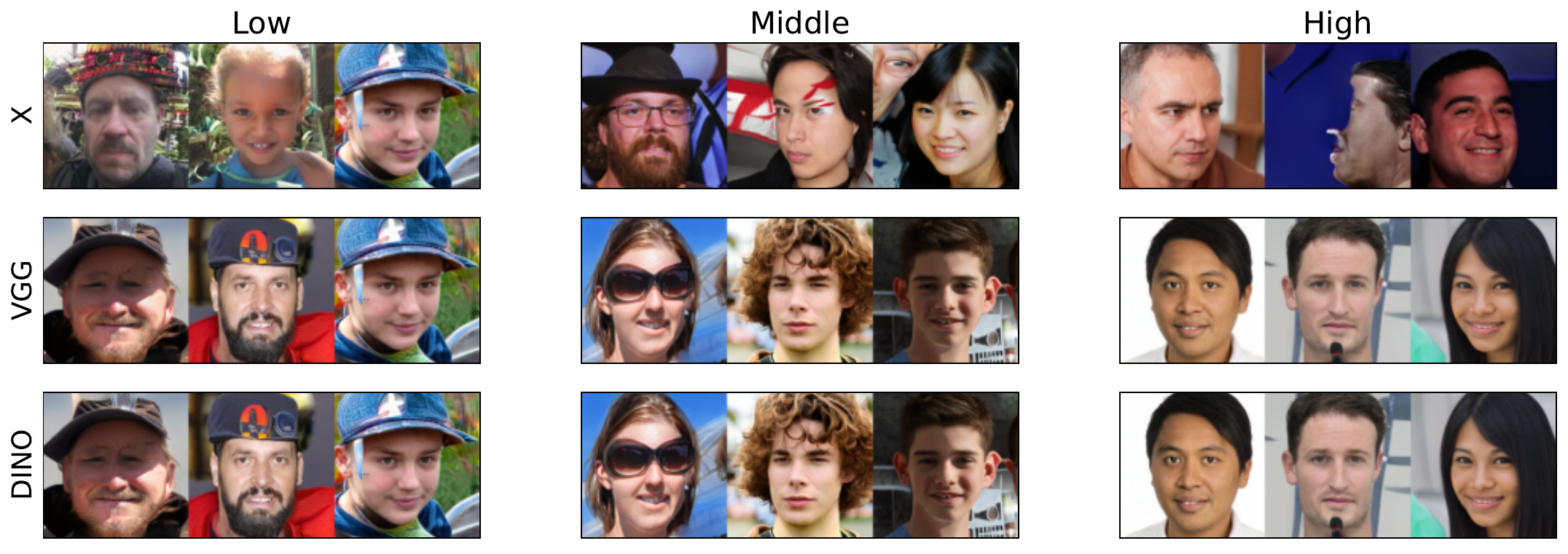}
\caption{Qualitative comparison of different image features. Three types of output considered: original image (X), VGG19 features (VGG), Dino features (DINO). For each type the sample density was computed. Three images with low, middle, and high scores are presented for each output type.}
\label{fig:features_comp}
\end{figure*}

\subsection{Feature Density Scoring}

High-resolution images can be generated by high-quality GANs (Generative Adversarial Networks). For instance, the StyleGAN2 model trained on the FFHQ dataset can generate images of size $1024\times1024$ with a latent space size of 512. However, computing the Jacobian matrix, which contains $10^9$ values in this case, can be challenging in terms of both time and memory consumption.

To mitigate these costs, we propose to use feature extraction. Instead of evaluating sample quality based on pixel values, a score function can be used that assesses the density of features. The score function is defined as:

$$s(x) = \log \rho(f(x)) = \log(\rho(z)) - \log(\mathrm{Vol}(J))$$
Here, $J = \frac{d f(G(z))}{dz}$ represents the Jacobian matrix, and $f$ denotes the feature extraction network. In Figure~\ref{fig:pipeline} we presented a whole pipeline of our work. The detailed explanations on last step delivered in next section.

In our work, we considered two options for the feature extraction network: VGG19\cite{vgg19} and Dino\cite{dino}. VGG networks are based on convolutional layers and have demonstrated high efficiency in classification tasks. They are widely used for extracting meaningful information from image data. On the other hand, Dino is a transformer-based network, which is known to be more accurate but has longer inference times compared to convolution-based networks.

To compare the performance of these feature extraction networks, we generated 100 images StyleGAN 2 ADA model trained on FFHQ dataset and evaluated the proposed metric. The experiments were conducted on a Tesla V100-SXM2 GPU with 16 GB of memory. Three types of output were considered: X (pixel-based density), VGG (VGG19-based feature density), and Dino (Dino-based feature density).

\begin{table}[t]
\centering
\begin{tabular}{c|ccc}
\hline
\textbf{\begin{tabular}[c]{@{}c@{}}Feature\\ extractor\end{tabular}}   & Original & VGG & Dino \\ \hline
\textbf{\begin{tabular}[c]{@{}c@{}}Time per\\ sample (s)\end{tabular}} & 450      & 25  & 90   \\ \hline
\end{tabular}
\caption{Comparison results of time consumption for metric inference. Three types of output density considered: Original - pixel based density, VGG - VGG features based density, Dino - Dino features based density}
\label{tab:time_features_comp}
\end{table}

After conducting a sorting procedure, we selected three images with the lowest, middle, and highest scores for each output type. Figure~\ref{fig:features_comp} illustrates that VGG produces results that are comparable to pixel-based and Dino based density. Additionally, computing the VGG features-based density requires significantly less time per sample. For further experiments, we utilized VGG features for metric computation. Table~\ref{tab:time_features_comp} provides a comprehensive comparison of the time consumption for different output types.

\subsection{Inference Time Acceleration through Tensor Train}
Presented in previous section reduction in inference time is insufficient for the computation of a large number of samples. Currently, the computation of 50,000 scores takes around 2 weeks on a single GPU, even with the use of VGG based features. A potential solution to this problem is to discretize the metric on a grid and compress the resulting tensor using tensor decomposition. The proposed solution pipeline consists of two stages:

\begin{enumerate}
    \item Score computation for a large number of samples
    \item Computation of the logarithm density approximation using the samples obtained from the previous stage
\end{enumerate}

To evaluate a sample $\hat{x}$ within this pipeline, the following steps can be followed:

\begin{enumerate}
    \item Find the closest point to the latent code $\hat{z}$ in the discrete latent space $z[i_1,...,i_d]$. This can be achieved by minimizing the Euclidean distance between the discrete latent codes and $\hat{z}$:
    $$(\hat{i}_1,...,\hat{i}d) = \arg\min{(i_1,...,i_d)}\|z[i_1,...,i_d] - \hat{z}\|$$
    \item Compute the score value at this point using the density tensor $\rho[i_1,...,i_d]$ stored in compressed format: 
    $$s(\hat{x}) = \rho[\hat{i}_1,...,\hat{i}_d]$$
\end{enumerate}

\subsection{Non uniform grid}

One crucial component in our scheme is the grid used for discretization. We found usage of a uniform grid is not suitable for GANs when sampling latent codes from a normal distribution. Certain latent regions may lack sufficient data, posing challenges for computation of tensor train descomposition.

To address this challenge, we opted for a grid where the integrals of the normal density over each grid interval are equal. This approach ensures a uniform distribution of samples along each grid index, effectively working in our case. More details can be found in Appendix A.

\begin{figure*}[t]
\centering
\includegraphics[width=0.9\textwidth]{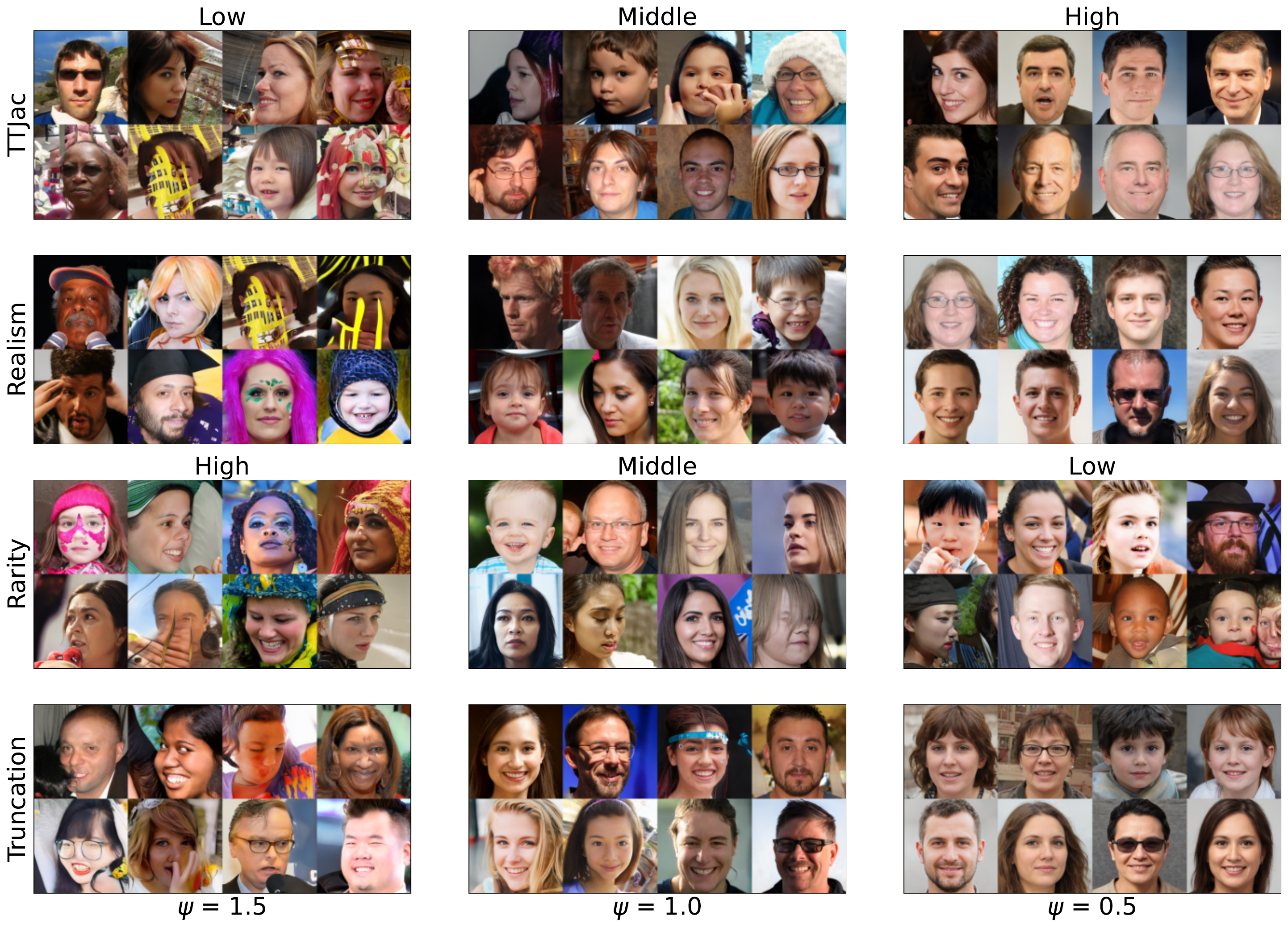}
\caption{Qualitative comparison of metrics for individual image evaluation: TTJac score, Realism score, Rarity score, Truncation Trick. For each metric we presented 8 images with lowest, middle, and highest score values. For the TTJac and Realism scores, we arranged the images in increasing order from low to high scores. Conversely, for the Rarity score and Truncation Trick, we intentionally reversed the order to ensure ease of comparison between the metrics.}
\label{fig:metrics_comp}
\end{figure*}

\begin{table}[t]
\centering
\begin{tabular}{c|cccc}
\hline
\textbf{Domain} & FFHQ  & Wild & Car & Horse \\ \hline
\textbf{MSE}    & 0.018 & 0.026     & 0.017    & 0.018      \\ \hline
\end{tabular}
\caption{Quantitative evaluation of TT approximation of TTJac score on discrete grid for different domains: FFHQ, AFHQ-Wild, LSUN Car, LSUN Horse}
\label{tab:quant_eval_approx}
\end{table}

\subsubsection{Calculation of Tensor Train approximation}

The TT decomposition of a tensor represents the element at position $[i_1, ..., i_N]$ as the product of matrices:

$$T[i_1, ..., i_N] = G_1[i_1]...G_N[i_N]$$

Here, $G_1, ..., G_N$ are the TT cores. The low-rank pairwise dependency between tensor components allows for an efficient approximation of the proposed metric function in discrete form using well known Tensor Train decomposition \cite{tt0}. It showed impressive results on different tasks \cite{tt1,tt2,tt3,tt4} due to the storage efficiency, capturing complex dependencies and fast inference capabilities. Obtaining an element of the tensor stored in TT format requires only $511$ matrix-by-vector multiplications. It also can be easily accelerated for the batch of elements using multiprocessing tools. Actually the batch size could be enormously large for TT format since it on each step of output computation algorithm stores the matrix of size $(N,r)$ where $N$ number of elements to compute, $r$ - rank of decomposition. This aspect has significat impact on use of proposed metric for truncation trick, where we need to evaluate huge number of samples. 

However, it should be noted that the computation of samples for core computation is time-consuming. This renders standard iterative methods for tensor train calculation ineffective, as they tend to overfit to given tensor samples and fail to capture underlined dependencies. In such case, it is more suitable to use explicit methods for tensor train decompositions, such as ANOVA decomposition \cite{anova}. The authors in \cite{tt1} have presented an effective method for computing ANOVA decomposition in TT format, which we have found accurate enough for our purposes. See Table~\ref{tab:quant_eval_approx} where we presented the results of TT approximation for discretized metric in different domains.

\subsubsection{Upgraded Truncation Trick}

The truncation trick, initially proposed in \cite{truncationtrick}, offers a means to adjust the balance between variability and fidelity. It involves two practical steps: evaluating the norm of generated samples and resampling those with a norm exceeding a specified threshold. This threshold determines the balance between variability and fidelity. By removing samples with high norms, we effectively reduce the number of samples with low latent code density and, possibly, lower visual quality. This approach can enhance the overall fidelity of the generated samples, albeit at the expense of variability.

Instead of evaluating latent code density, we suggest assessing image feature density using a modern feature extractor such as VGG. In next section, we provide evidence that this replacement criterion achieves a more desirable trade-off between fidelity and variability.

\section{Experiment}

To evaluate the proposed metric, we conducted experiments on various widely-used datasets for image generation, including FFHQ \cite{ivgeneration3}, AFHQ-Wild \cite{translation4}, LSUN-Cars \cite{lsun}, and LSUN-Horse \cite{lsun}. Additionally, we considered the method effectiveness on StyleGAN 2 \cite{ivgeneration4} and StyleGAN 2 ADA \cite{ivgeneration5} models, known for their high performance in generating images in standard domains where data may be limited or noisy.

For the learning process of the metric, we computed $60k$ samples for the FFHQ dataset and $30k$ samples for the AFHQ-Wild, LSUN-Cars, and LSUN-Horse datasets. Feature extraction was performed using the VGG19 network \cite{vgg19}.

To discretize the metric, we utilized a grid with a size of $32$. The metric was then approximated by applying ANOVA decomposition of order 1, which was later converted to the tensor train format.

All experiments were carried out using a 3 Tesla V100-SXM2 GPU with 16 GB of memory.

\begin{figure}[t]
\centering
\includegraphics[width=0.45\textwidth]{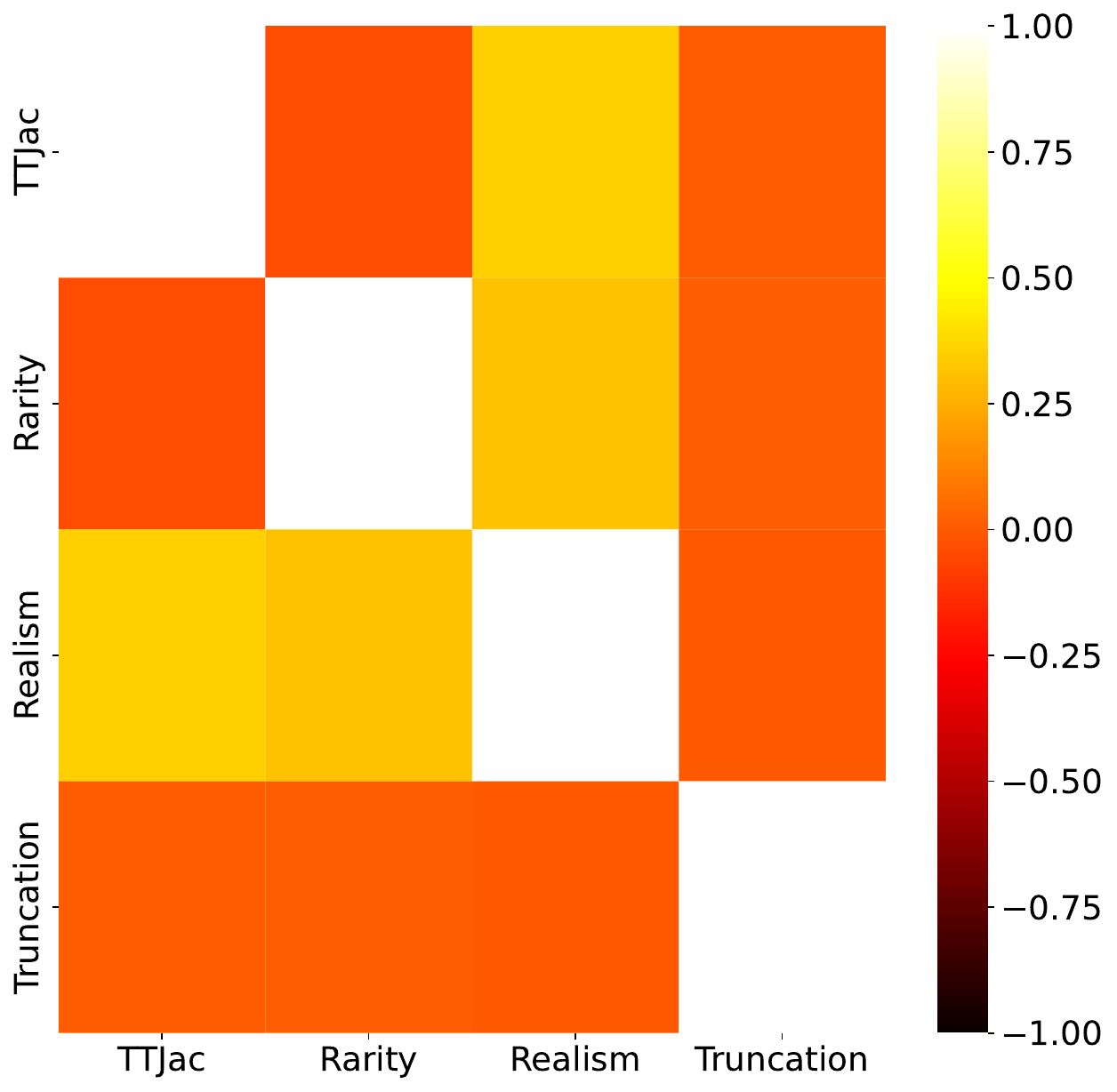}
\caption{Correlation matrix for metrics measuring individual sample quality: TTJac score, Realism score, Rarity score, and Truncation.}
\label{fig:metrics_corr}
\end{figure}

\begin{table}[t]
\centering
\begin{tabular}{c|ccc}
\hline
\textbf{\begin{tabular}[c]{@{}c@{}}Number\\ of border\\ samples\end{tabular}} & \textbf{Rarity} & \textbf{Realism} & \textbf{Truncation} \\ \hline
3000                                                                          & -0.051           & 0.42            & 0.007               \\
2000                                                                          & -0.059          & 0.472            & 0.156               \\
1000                                                                          & -0.048          & 0.54            & 0.006               \\
500                                                                           & -0.048          & 0.584            & 0.049               \\
100                                                                           & -0.043          & 0.651            & 0.135               \\ \hline
\end{tabular}
\caption{Quantitative results of correlation between TTJac score and other considered metrics: Rarity score, Realism score, Truncation. Computation was done for certain number of samples representing the higher and lower extremes of the TTJac score - border samples.}
\label{tab:corr_comp_metrics}
\end{table}

\subsection{Comparison with other metrics}

To demonstrate the effectiveness of the TTJac score in evaluating individual samples, we compared it with similar metrics such as the Realism score \cite{precisionrecall}, Truncation trick \cite{truncationtrick}, and Rarity score \cite{rarity}. We randomly selected $10k$ latent samples and presented the comparison results in Figure~\ref{fig:metrics_comp}.

In order to facilitate the comparison process, we reversed the order for the Rarity score and Truncation. The Realism score showed high results in evaluating sample fidelity but sometimes failed on visually appealing images. It is important to note that high realism values often correspond to images with low variability. Similarly, the Truncation trick allows for manipulation of image quality at the expense of variability. However, samples with the highest variability tend to have lower realism compared to other metrics. On the other hand, the Rarity score measures the uniqueness of the given image, effectively identifying images with distinct features, but some of them may appear less realistic.

The proposed TTJac score functions similarly to the realism score but does not require any real data for processing. It effectively extracts samples with high fidelity, reflected by a high score. Conversely, low scores often indicate the presence of visual artifacts. However, the TTJac score does have a limitation – images with high scores tend to have fewer unique features, while visually appealing images can be found among those with low scores.

We also computed the correlation between the presented metrics. The results confirmed quite high similarity between the realism and TTJac scores, as shown in Figure~\ref{fig:metrics_corr}. Furthermore, when measuring metrics correlation for samples with the highest and lowest TTJac scores, results of proposed metric becomes even more close to realism score. See Table~\ref{tab:corr_comp_metrics} for confirmation. Thus, the data free metric TTJac is able to filter out very low quality images as effectively as the Realism score using a dataset of real images.

\begin{figure*}[h!]
\centering
\includegraphics[width=0.9\textwidth]{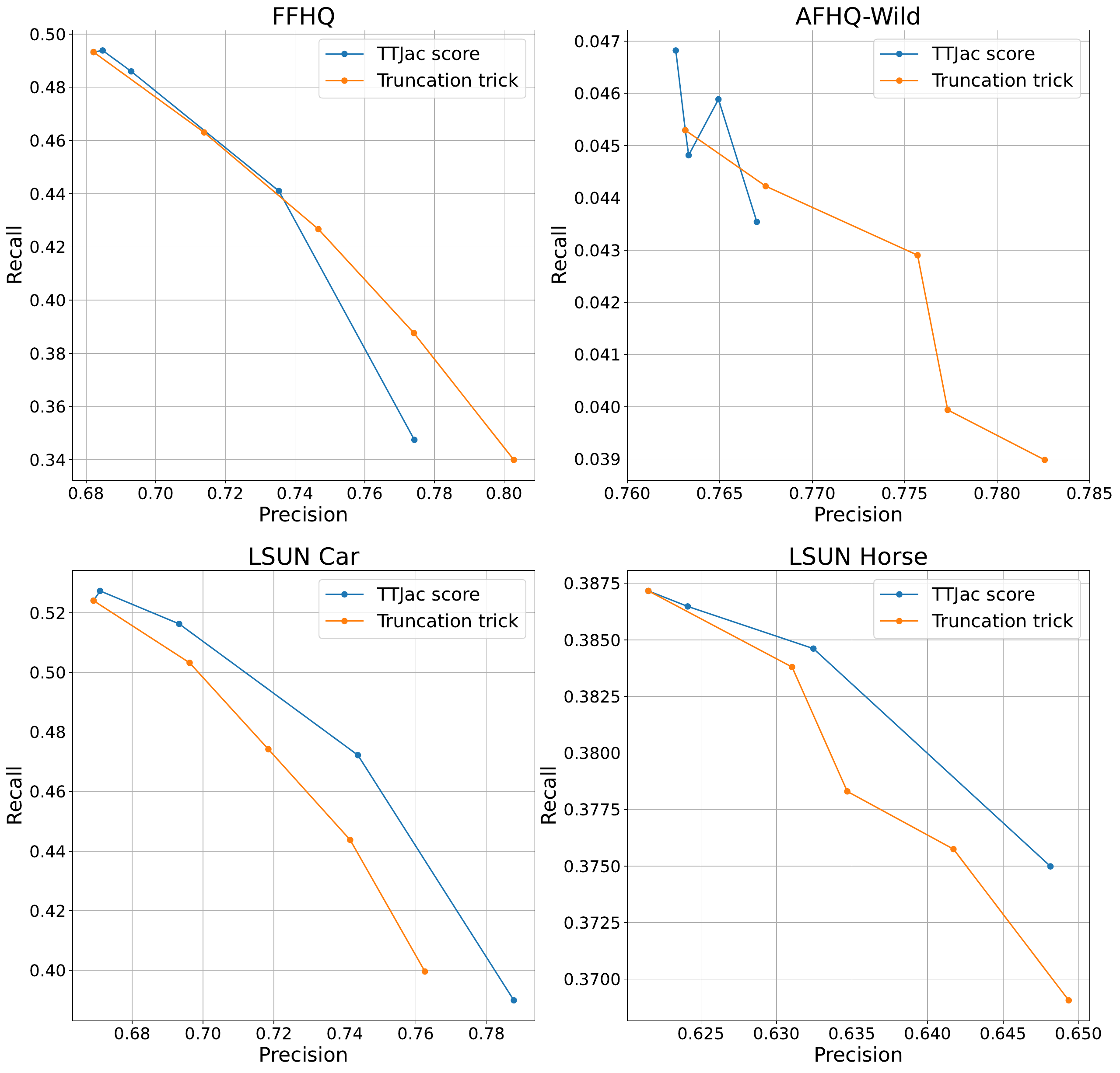}
\caption{Quantitative comparison of fidelity-variability trade-off computed using TTJac score and Truncation trick. Four domains were examined: FFHQ, AFHQ-Wild, LSUN Car, LSUN Horse. For each domain $50k$ samples were generated for precision and recall calculation. The results were averaged along 3 random seeds. The higher - the better.}
\label{fig:balance_comp}
\end{figure*}

\begin{figure*}[t]
\centering
\includegraphics[width=0.9\textwidth]{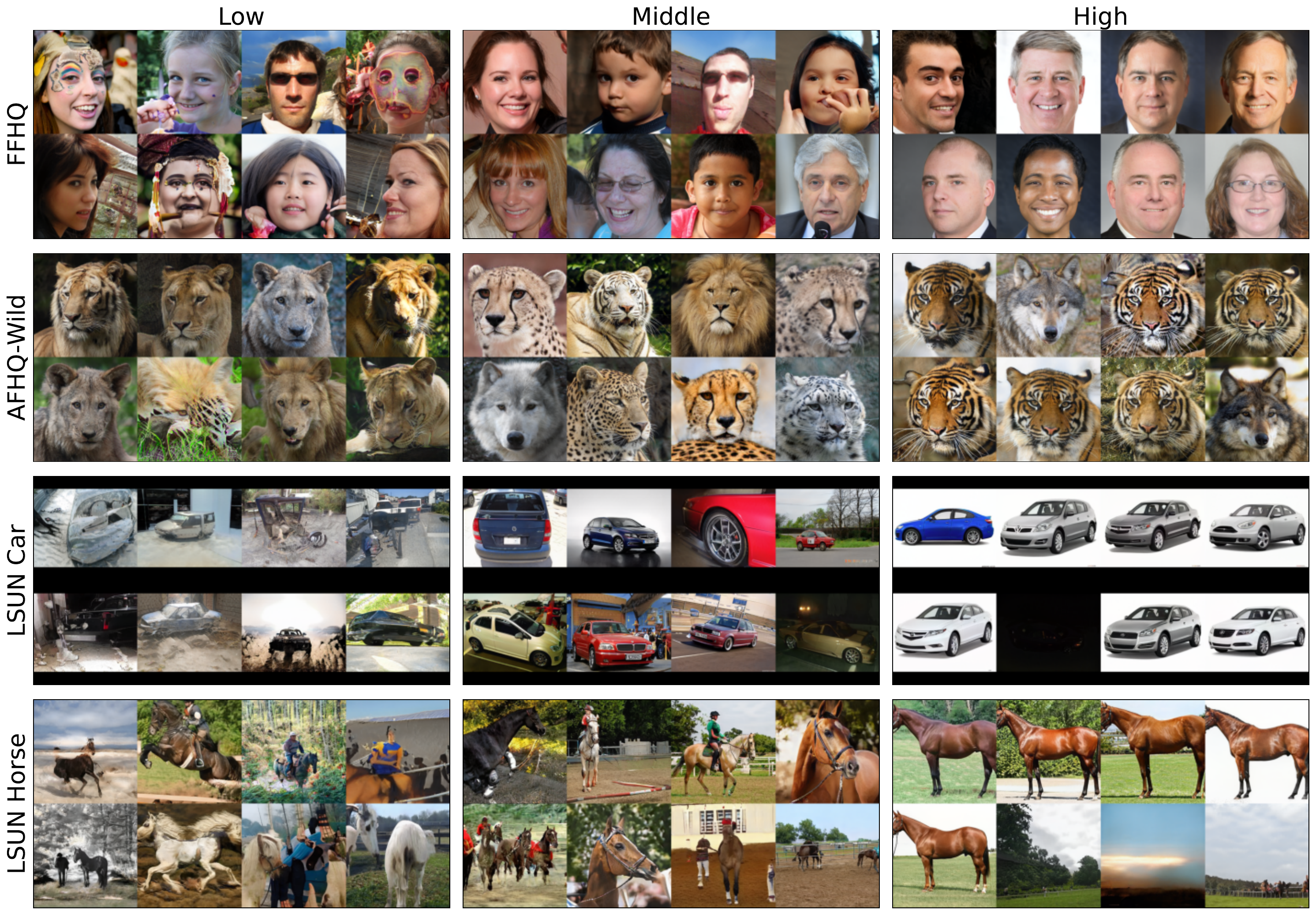}
\caption{Qualitative comparison of metric evaluation capabilities four domains were examined: FFHQ, AFHQ-Wild, LSUN Car, LSUN Horse. For each domain $30k$ samples were sorted based on their scores and selected three images to represent the samples with the lowest, middle, and highest score values.}
\label{fig:full_img_comp}
\end{figure*}

\subsection{Fidelity-variability trade-off evaluation}

In method section, we discussed the potential use of the proposed metric to enhance the performance of the truncation trick. In this section, we compare the trade-off provided by the standard criteria based on latent code norm, and the TTJac score. To demonstrate the effectiveness, we plotted precision-recall curves for the FFHQ, AFHQ-Wild, LSUN-Car, and LSUN-Horse domains.

For the FFHQ, LSUN Horse, and LSUN Car domains, the use of the TTJac score allows for a better balance between precision and recall compared to the standard tool. In Figure~\ref{fig:balance_comp}, the curve for the TTJac score consistently lies above the curve for the Truncation trick. The benefit is not significant for the AFHQ-Wild domain. This can be attributed to the fact that the GAN model already exhibits a high level of precision in this specific domain. Furthermore, the evaluation of TT approximation accuracy presented the least favorable outcome when compared to other domains. And also on the FFHQ domain, after an improvement in precision by 5\%, a decline begins and the curve falls below that for the standard truncation trick. It can be attributed to the presence of an error in the metric approximation. However, for the LSUN-Car domain, the TTJac score proves to be highly effective, providing a significant improvement in precision with negligible loss in recall.

It is important to note that due to the presence of errors in the approximation of the metric score, it is not possible to achieve the maximum possible precision with minimal recall value, as seen in standard precision-recall curves.

Overall, this comparison highlights the potential of the TTJac score in achieving a better trade-off between precision and recall in the evaluated domains, with notable improvements observed in the LSUN-Car domain.

\subsection{Domain wise metric evaluation}

In this part, we conducted a qualitative evaluation of the TTJac score for various domains. In Figure~\ref{fig:full_img_comp}, it is demonstrated that the TTJac score efficiently detects visual artifacts in the evaluated domains.

For the FFHQ domain, the TTJac score assigns low scores to images that do not contain a face or have unrealistic prints. Additionally, images with visually poor backgrounds are also marked with low score values. Similarly, in the LSUN Car domain, images that lack key elements of a car are identified as lower quality by the TTJac score.

The TTJac score exhibits similar behavior in the AFHQ-Wild and LSUN-Horse domains. It effectively detects unrealistic horses, such as instances where two horses are merged into one image. However, in the AFHQ-Wild domain, the metric faces a more challenging situation. While it accurately identifies images where different species are merged, such as an image with half wolf and half lion, the overall fidelity of the images in this domain is quite close. This observation is connected to the fact that the model trained on the AFHQ-Wild dataset has a higher precision compared to the other considered models.

In summary, the TTJac score demonstrates high efficiency in evaluating sample fidelity while sacrificing some variability. It effectively detects visual artifacts and accurately identifies unrealistic elements in the evaluated domains.

\section{Conclusion}
We have proposed a new approach for evaluating image quality without using real data, which is based on the density of meaningful features extracted from the image. A method was also proposed for efficient storage and inference metrics using TT approximation. We compared the TTJac score with other metrics. The TTJac score performs similarly to the realism score. It effectively detects visual artifacts and identifies unrealistic elements in different domains such as FFHQ, AFHQ-Wild, LSUN Car, and LSUN Horse. We also evaluated the trade-off between fidelity and variability using precision-recall curves. The TTJac score showed a better balance, especially in the LSUN Car domain where it significantly improved precision with minimal loss in recall. In qualitative evaluation, the TTJac score successfully detected missing key elements or unrealistic features in images across various domains. Overall, the TTJac score demonstrates high efficiency in evaluating sample fidelity and can be a valuable tool for assessing image generation models.

\bibliography{main}
\newpage
.
\appendix
\newpage
\section{Appendix A} 
The general representation of a tensor $T$ in the TT format is:

$$T[i_1,...,i_d] \approx G_1[i_1] G_2[i_2] ... G_d[i_d]$$

Here, $G_1,...,G_d$ are matrices of size $(r,r)$ (if the compression rank is equal), and $d$ is the size of the latent space.

The most common way to find the cores $G_k$ for a tensor represented by random elements is the alternative least squares algorithm (ALS) \cite{als}. This algorithm consists of an alternating best fit for each of the cores $G_k$ in the order of $k=1,...,d$. For each core $G_k$, we need to solve an independent least squares problem:

$$\hat{G}k = \arg\min{G_k}|T - [G_1,...,G_k,...G_d]|$$

Here, $[G_1,...,G_k,...G_d]$ is a short notation for the tensor represented in the TT format, and $\hat{G}_k$ is the solution for the least squares problem. This problem can be solved independently for each matrix $G_k[i]$, where $i = 1, ..., N$ (N is the grid size):

$$\hat{G}_k[i] = \arg\min{G_k[i]}\|T^{k}_i - L_k G_k[i] R_k \|$$

In the above equation, $L_k[i_1,...,i_{k-1}] = G_1[i_1]...G_{k-1}[i_{k-1}]$ left side of TT format for core $k$, $R_k[i_{k+1},...,i_{d}] = G_{k+1}[i_{k+1}]...G_{d}[i_{d}]$ right side of TT format for core $k$, and $T^k_i[i_1,...,i_{k-1},i_{k+1},...,i_d] = T[i_1,...,i_k=i,...,i_d]$ represents a $(d-1)$-dimensional target tensor. 

If we have $M$ tensor elements, then $L_k$ becomes a set of vectors $L_k[m] = L_k[i_1=i_1^m,...,i_{k-1}=i_{k-1}^m]$ representing the left side of the TT format for $m=1,...,M$ samples, $i_1^m,...,i_{d}^m$ - indexes of $m$-th tensor sample. $R_k$ similarly turns into a corresponding set of vectors $R_k[m] = R_k[i_{k+1}=i_{k+1}^m,...,i_{d}=i_{d}^m]$. Both $L_k$ and $R_k$ here have size $(M_i^k,r)$, where $M_i^k$ represents the number of samples for which $i_k = i$. In this case, $T^k_i$ becomes a vector of values of tensor samples corresponding to $i_k = i$. Finally the problem can then be formulated as a standard linear problem:

$$\hat{G}_k[i] = \arg\min{G_k[i]}\|T^{k}_i - (R_k \odot L_k) \mathrm{vec}(G_k[i])\|$$

In the above equation, $R_k \odot L_k$ represents the face-splitting product of matrices $L_k$ and $R_k$ of size $(M_i^k,r^2)$, and $\mathrm{vec}(G_k[i])$ represents the vectorization of matrix $G_k[i]$. To solve the problem, $M_i^k$ should be greater than or equal to $r^2$ for any $k$ and $i$. If this condition is not met, the system becomes underdetermined, necessitating the setting of additional constraints.

This is the main problem where the grid type affects tensor decomposition. Let us consider a uniform grid with the following parameters: minimal value $a=-3$, maximal value $b=3$, and grid size $N=64$. For a normally distributed variable $z_k$, the relative number of samples inside interval $[t_i,t_{i+1}]$ is given by:

$$p^k_i = \frac{\int_{t_i}^{t_{i+1}}\exp(-z_k^2/2)}{\int_{t_0}^{t_{N-1}}\exp(-z_k^2/2)}$$

Here, $t_i = a + \frac{(b-a)}{(N-1)}i$, and $i$ represents the grid index in the range $[0,N-1]$.

Based on this, we can find the ratio between the number of samples inside the first interval $i=0$ and the middle one $i=N/2-1=31$:

$$\frac{p^k_0}{p^k_{31}} = 0.013$$

The discrepancy in the number of equations between $i=0$ and $i=31$ poses a limitation on the rank of the decomposition. For $i=31$, the system has approximately 100 times more equations compared to $i=0$. Additionally, when considering the total number of samples $M=50k$, on average, only 24 samples have $i=0$ restricting the rank of the decomposition to a maximum of 4.

To address this issue, we propose using a non-uniform grid with intervals $[t_0,t_1],[t_1,t_2],...,[t_{N-2},t_{N-1}]$ such that the integral of the normal density over these intervals is equal:

$$\int_{t_i}^{t_{i+1}}\exp\left(-\frac{z_k^2}{2}\right) = \int_{t_j}^{t_{j+1}}\exp\left(-\frac{z_k^2}{2}\right)$$

Here, $i$ and $j$ can be any arbitrary values in the range $[0,N-2]$. By imposing this condition on grid points, the number of equations in the linear system becomes equal for all values of $i$.

\begin{figure}[t]
\centering
\includegraphics[width=0.45\textwidth]{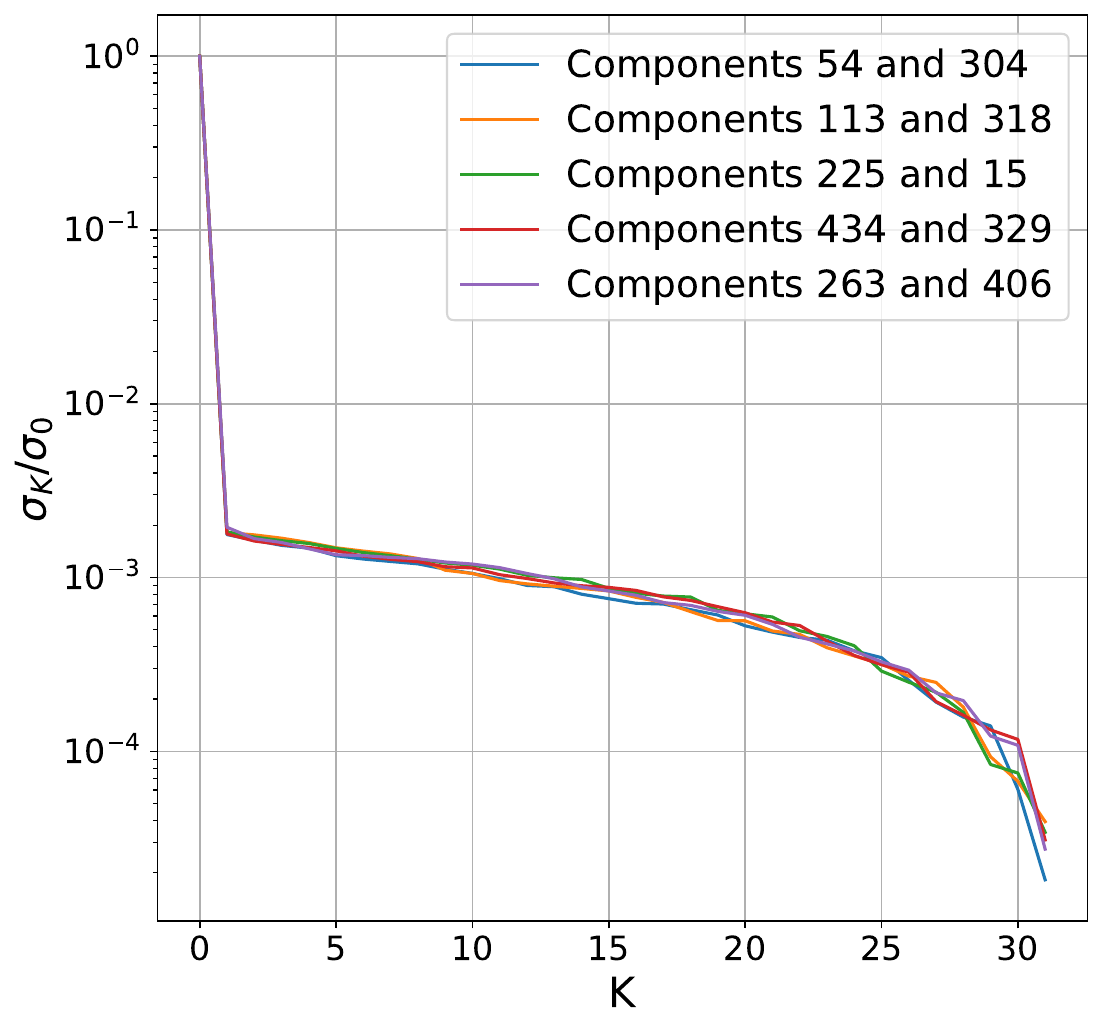}
\caption{Normalized cumulative sum of singular values for matrix $C$ computed for random pairs of components}
\label{fig:dependence_components}
\end{figure}

\section{Appendix B}
\label{sec:appb}
Computation of tensor train cores rise a question of choosing optimal hyperparameters like rank, method, core initialization, number of iterations (if applicable). To find the optimal rank we constructed the matrix $C$ describing the pair wise dependency between two components $k_1$ and $k_2$:

$$C[j_r,j_c] = \mathop{\mathbb{E}}\limits_{i_k, k\neq [k_1, k_2]}T[i_1,...,i_{k_1}=j_r,...,i_{k_2}=j_c,...,i_d]$$

where $C[j_r,j_c]$ - the element on intersection of $j_r$-th row and $j_c$-th column. Taking into consideration the TT approximation $T \approx [G_1,...,G_d]$ we can express previous equation from TT format side:

\[C[j_r,j_c] \approx \mathop{\mathbb{E}}\limits_{i_k, k\neq [k_1, k_2]}G_1[i_1]...G_{k_1}[i_{k_1}=j_r]...\\ G_{k_2}[i_{k_2}=j_c]...G_d[i_d]\]

Here $G_1$ is a vector (first TT core). Summation made along second dimension of each core produces the following result:

$$C[j_r,j_c] \approx G_1^{k_1-1}G_{k_1}[i_{k_1}=j_r]G_{k_1+1}^{k_2-1}G_{k_2}[i_{k_2}=j_c]G_{k_2+1}^d$$

where $G_1^{k_1-1}$ is a vector representing product of averaged along middle dimension cores before $k_1$ component, $G_{k_1+1}^{k_2-1}$ is a matrix of size $(r,r)$ representing product of averaged along middle dimension cores after $k_1$ and before $k_2$ components, $G_{k_2+1}^d$ is a vector representing product of averaged along middle dimension cores after $k_2$ component.

The rank of matrix $C$ is equal to $\mathrm{rank}(G_1^{k_1-1}G_{k_1}G_{k_1+1}^{k_2-1}G_{k_2}G_{k_2+1}^d) \leq r$ since the smallest dimension size of tensors in the product is $r$. So following this formulation we could estimate the optimal rank by evaluating the rank of matrix $C$ computed for target tensor $T$. Figure~\ref{fig:dependence_components} presents normalized singular values for matrix $C$ computed for different random pairs of tensor components. Based on results we can conclude that tensor could be accurately compressed with small rank.

\end{document}